\documentclass[conference]{IEEEtran}
\IEEEoverridecommandlockouts

\usepackage{cite}
\usepackage{amsmath,amssymb,amsfonts}
\usepackage{graphicx}
\usepackage{textcomp}
\usepackage{xcolor}

\usepackage{algorithm}
\usepackage{algpseudocode}

\usepackage{amsmath,amssymb}

\usepackage{cite}
\usepackage{url}
\usepackage[hidelinks]{hyperref}

\usepackage{xcolor}
\usepackage{microtype}

\usepackage{booktabs}
\usepackage{multirow}
\usepackage{array}
\usepackage{makecell}
\usepackage{tabularx}
\usepackage{siunitx}

\def\BibTeX{{\rm B\kern-.05em{\sc i\kern-.025em b}\kern-.08em
    T\kern-.1667em\lower.7ex\hbox{E}\kern-.125emX}}
\begin{document}

\title{TSCA-Net: Temporal-Spatial Clique Attention for Interpretable Multimodal Pedestrian Trajectory Prediction}

\author{
\IEEEauthorblockN{
Md Mustafizur Rahman\textsuperscript{1},~
Guangchao Yang\textsuperscript{1,*},~
A F M Abdun Noor\textsuperscript{2},~
Md Imam Ahasan\textsuperscript{1}, \\
Md Mahfuzur Rahman\textsuperscript{1},~
Md Ariful Islam\textsuperscript{1}
}

\IEEEauthorblockA{
\textsuperscript{1}College of Computer Science, Chongqing University, Chongqing, China\\
\textsuperscript{2}Department of Software Engineering, Daffodil International University, Dhaka, Bangladesh
}

\thanks{Corresponding author: gchao\_yang@cqu.edu.cn. This work was supported by the National Natural Science Foundation of China under Grant 62572086.}
}


\maketitle

\begin{abstract}
Accurate pedestrian trajectory prediction in crowded environments remains challenging due to the multimodal uncertainty of human motion and the variable complexity of motion dynamics across different scene contexts. Existing goal-conditioned models rely on static displacement structures that assign equal weight to all historical time steps, standard graph attention mechanisms, and fixed-capacity motion decoders that cannot adapt to local prediction complexity. To address these limitations, we propose TSCA-Net, a trajectory prediction framework built upon three complementary modules. The Temporal-Spatial Clique Attention (TSCA) module introduces learnable temporal gating into clique-based goal-history interaction, enabling time-aware modulation of historical observations relative to each candidate goal. The Cross-Pedestrian Clique Potential (CPCP) module models asymmetric pairwise agent relationships through a dynamic clique potential framework with a time-varying social graph. The Adaptive KAN Grid Refinement (AKGR) mechanism dynamically adjusts the B-spline grid resolution of a Kolmogorov-Arnold Network-augmented LSTM decoder based on per-agent goal distribution entropy, balancing model expressiveness against overfitting across varying motion complexities. Extensive experiments on the ETH/UCY and Stanford Drone Dataset benchmarks demonstrate that TSCA-Net achieves very good state-of-the-art performance, an average ADE/FDE of 0.13/0.20\,m on ETH/UCY and 6.95/10.43 pixels on SDD, representing improvements of 13.3\% and 5.96\% in ADE over the preceding MCK-Net baseline, respectively. Comprehensive ablation studies confirm that all three modules contribute complementary and mutually reinforcing improvements. Code is available at: https://github.com/imamahasane/TSCA-Net.
\end{abstract}


\section{Introduction}

Predicting the future trajectories of individuals in crowded or roadside areas is a critical task, particularly for applications such as social robots~\cite{baillie2019challenges}, autonomous driving~\cite{bai2015intention, luo2018porca, muhammad2020deep}, and video surveillance~\cite{nguyen2019anomaly, gong2021enhanced}. However, this task is challenging due to the involvement of human perception, motion analysis, and reasoning. It requires simultaneous consideration of each pedestrian's intended destination and the impact of surrounding pedestrians on the focal pedestrian. Anticipating pedestrian trajectories is therefore crucial for systems requiring active environmental response and precise pedestrian localization. Research on pedestrian trajectory forecasting has long been conducted, and a common finding across most studies is that pedestrians tend to plan their behaviors under specific social norms~\cite{chen2023vnagt, zhou2022spatiotemporal, gupta2018social, alahi2016social, mohamed2020social, shi2021sgcn, cai2021environment}. Pedestrians moving toward one another anticipate potential collision points based on their walking speed and plan their movements to avoid predictable conflicts. In other words, pedestrians sharing the same scene typically form implicitly interacting relationships under normal circumstances. To capture these social dynamics, deep learning-based methods have been widely adopted and have achieved impressive results in pedestrian trajectory prediction, prompting the proposal of numerous network architectures for this task~\cite{li2024pedestrian, liu2024modeling, Salzmann_2020_ECCV, jain2020discrete}. These models learn from observed motion history while accounting for various factors, including social influences and individual behaviors, to forecast future trajectories.

One straightforward approach is to incorporate contextual factors such as the social environment and motion interactions during prediction~\cite{peng2023mrgtraj, wang2023trajectory, yang2023long}. However, effectively modeling these factors and integrating them coherently into a network remains difficult. Methods based on scene information~\cite{feng2024multi, liang2020garden, 2021Safety} treat the scene holistically and leverage models to learn the latent associations among factors in the scene environment, avoiding the need to model each factor independently and achieving notable success. Building on these foundations, various network frameworks have been proposed to enhance prediction performance, including generative model-based approaches~\cite{2023Reciprocal, Sadeghian_2019_CVPR}, temporal-based structures~\cite{Yuan_2021_ICCV, guo2022, 9636722}, and transformer-based architectures~\cite{10367756}. These models address different aspects of the prediction problem from multiple perspectives, making significant contributions to overall performance. Nevertheless, accurately predicting pedestrian trajectories in complex and dynamic open environments remains a fundamentally challenging task, primarily due to the inherent multimodality and uncertainty of human motion, which has driven researchers toward incorporating richer contextual and social cues into their models.

Recent works enhance the understanding of road-agent behavior by introducing physical constraints and social interactions, such as considering context information (map information, spatial information, system dynamics, etc.)~\cite{tang_etal_2024_itinera, Lee_2017_CVPR, Sadeghian_2019_CVPR, Marchetti_2020_CVPR} and modeling social interaction behaviors between agents during trajectory prediction~\cite{Mohamed_2020_CVPR, NEURIPS2019_d09bf415}. While emerging models continually emphasize social interactions, Makansi et al.~\cite{Makansietal21} and Saadatnejad et al.~\cite{SAADATNEJAD2022103705} questioned the true contribution of social interaction mechanisms to trajectory prediction accuracy. Experimental results in~\cite{Makansietal21} further demonstrate that existing state-of-the-art models~\cite{Mohamed_2020_CVPR, Salzmann_2020_ECCV, Mangalam_2020} exhibit similar prediction performance even after removing interaction features via feature attribution methods. In contrast to a pure context-information perspective, several methods emphasize the importance of understanding agents' goals when predicting future trajectories~\cite{Mangalam_2020, Zhao_pmlr_2021}. PECNet conditions predictions on the estimated final location of the future trajectory to obtain more interpretable and precise predictions~\cite{Mangalam_2020}. However, most of these works~\cite{Mangalam_2020, Choi_pmlr_2021, 9691856} are stochastic and leverage estimated goal states as precise conditioning signals, which limits their practicability and applicability in real-world settings.

A key limitation of existing goal-conditioned trajectory prediction models is that the interaction between candidate goals and the observed history is captured through a \emph{static} displacement structure, which assigns equal weight to all historical time steps regardless of their temporal proximity or relevance to the predicted endpoint. Moreover, standard social interaction modules treat agent neighborhoods as fixed graph structures, failing to capture the directional, asymmetric nature of pedestrian avoidance and group formation dynamics. Finally, conventional motion decoders apply a fixed nonlinear transformation to the hidden state regardless of the local complexity of the prediction problem, limiting their capacity to capture fine-grained behavioral modes in high-ambiguity regions such as intersections or narrow corridors.

To address these limitations, we propose \textbf{TSCA-Net}, a novel pedestrian trajectory prediction framework built on a hierarchical predict-then-refine paradigm. TSCA-Net introduces three complementary components atop the MCK-Net backbone~\cite{11157918}: a TSCA module for time-aware goal-history interaction, a Cross-Pedestrian Clique Potential (CPCP) module for asymmetric social interaction modeling, and an Adaptive KAN Grid Refinement (AKGR) mechanism for uncertainty-driven motion decoding. The main contributions of this work are summarized as follows:

\begin{itemize}
    \item We propose a novel attention block that introduces learnable temporal gating into clique-based goal-history interaction, enabling the model to selectively weight historical observations by their relevance to each candidate goal and producing temporally coherent goal distributions.

    \item We introduce a dynamic social interaction module that models asymmetric pairwise agent relationships via a clique potential framework with a time-varying social graph, more faithfully capturing pedestrian avoidance and group formation than symmetric graph attention.

    \item We present an entropy-driven mechanism that dynamically adjusts the B-spline grid resolution of a KAN-augmented LSTM decoder based on per-agent goal uncertainty, balancing model expressiveness against overfitting across varying motion complexity.
\end{itemize}

\begin{figure*}[!ht]
    \centering
    \includegraphics[width=\textwidth]{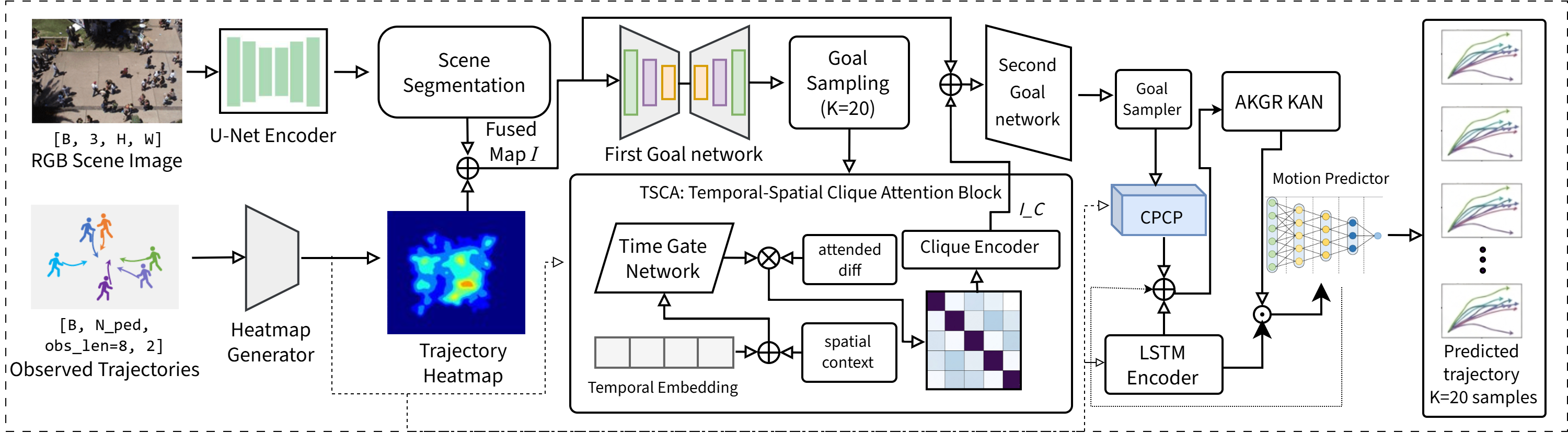}
    \caption{Overview of TSCA-Net. A dual-stream input module fuses scene and trajectory features into a unified map $\mathbf{I}$. The TSCA-Enhanced Goal Module produces an initial heatmap $\mathbf{M}^{(1)}$, refined via Temporal-Spatial Clique Attention into $\mathbf{M}^{(2)}$. The CPCP-AKGR Prediction Module decodes $K{=}20$ goal samples into full multimodal trajectories.}
    \label{fig:overall_model}
\end{figure*}

\section{Methodology}
\label{sec:methodology}

\subsection{Overall Architecture}
\label{subsec:overall_arch}

The overall architecture of TSCA-Net, illustrated in Fig~\ref{fig:overall_model}, follows a hierarchical predict-then-refine paradigm consisting of three sequential stages: (i) a dual-stream \emph{Input Module} that fuses scene-level and agent-level cues into a unified spatial representation; (ii) a \emph{TSCA-Enhanced Goal Module} that estimates a multi-modal goal distribution over the scene in two cascaded stages, where the second stage is conditioned on temporally-gated clique attention features; and (iii) a \emph{CPCP-AKGR Prediction Module} that iteratively decodes full trajectory paths by coupling social potential fields with an adaptive KAN-based motion model. This architecture extends the MCK-Net backbone~\cite{11157918} by replacing static interaction representations with dynamic, time-aware clique structures, and by introducing entropy-driven grid adaptivity into the prediction decoder.

\subsection{Input Module}
\label{subsec:input_module}

Accurate trajectory forecasting demands a joint understanding of two fundamentally distinct modalities: the \emph{static} topological layout of the environment and the \emph{dynamic} kinematic history of each agent. To this end, the input module processes both signals in parallel before fusing them into a shared spatial tensor.

\textbf{Scene Segmentation Encoder:} The RGB scene image $\mathcal{S} \in \mathbb{R}^{3 \times H \times W}$ is passed through a U-Net-style convolutional encoder~\cite{ronneberger2015unet} that produces a semantic segmentation map $\mathbf{F}_s \in \mathbb{R}^{C_s \times H \times W}$, where $C_s$ encodes pixel-wise semantic categories (walkable surface, obstacles, static structures). The use of a fully convolutional architecture with skip connections preserves fine-grained spatial resolution, ensuring that the model retains both local texture and global layout information simultaneously.

\textbf{Trajectory Heatmap Generator:} For each pedestrian $i$, the $T_{\text{obs}}$ observed positions are rendered onto a spatial grid of the same resolution $H \times W$ by placing Gaussian kernels centered at each coordinate, yielding an agent-specific occupancy heatmap $\mathbf{H}^{(i)} \in \mathbb{R}^{1 \times H \times W}$. Gaussian rendering is preferred over sparse positional encoding because it encodes spatial uncertainty implicitly and produces smooth gradients for downstream feature extraction.

\textbf{Fusion Map:} The scene segmentation $\mathbf{F}_s$ and the pedestrian heatmap $\mathbf{H}^{(i)}$ are concatenated channel-wise to produce the per-agent fused map:
\begin{equation}
    \mathbf{I}^{(i)} = \left[\mathbf{F}_s \,\|\, \mathbf{H}^{(i)}\right] \in \mathbb{R}^{(C_s + 1) \times H \times W},
    \label{eq:fused_map}
\end{equation}
where $[\,\cdot \| \cdot\,]$ denotes channel-wise concatenation. This representation $\mathbf{I}^{(i)}$ encodes both where the agent \emph{has been} and what environmental affordances are \emph{structurally permissible}, providing the goal and prediction modules with a spatially coherent conditioning signal.

\subsection{TSCA-Enhanced Goal Module}
\label{subsec:goal_module}

The goal module addresses the multi-modality of human navigation by formulating destination prediction as a probabilistic spatial distribution estimation problem over the scene, rather than regressing a single deterministic endpoint. It operates in two cascaded stages: an initial goal network that produces a coarse probability heatmap over plausible destinations, followed by a TSCA-conditioned second goal network that refines this distribution with temporally-grounded social context.

\subsubsection{First Goal Network}
\label{subsubsec:first_goal}

The first-stage goal network $\mathcal{G}^{(1)}$ processes the fused map $\mathbf{I}^{(i)}$ through a U-Net-structured encoder-decoder:
\begin{equation}
    \mathbf{M}^{(1)}_i = \sigma\!\left(\mathcal{G}^{(1)}\!\left(\mathbf{I}^{(i)}\right)\right) \in [0,1]^{H \times W},
    \label{eq:first_goal}
\end{equation}
where $\sigma(\cdot)$ denotes the sigmoid activation and $\mathbf{M}^{(1)}_i$ is interpreted as a probability map over all spatial locations being the final destination of agent $i$. The network is trained with a binary cross-entropy loss against a ground-truth goal heatmap constructed by placing a narrow Gaussian at the true final position $\left(x_{T_{\text{pred}}}^{(i)}, y_{T_{\text{pred}}}^{(i)}\right)$.
From $\mathbf{M}^{(1)}_i$, we sample $K{=}20$ candidate goal positions $\mathcal{G}_{K}^{(i)} = \left\{g_k^{(i)}\right\}_{k=1}^{K}$ via stratified spatial sampling, yielding a diverse set of future endpoint hypotheses that span the principal modes of the predicted distribution.

\subsubsection{Temporal-Spatial Clique Attention (TSCA)}
\label{subsubsec:tsca}

A core limitation of prior goal-conditioned trajectory models is that the interaction between candidate goals and the observed history is captured through a \emph{static} displacement structure, which assigns equal epistemic weight to all historical time steps regardless of their temporal proximity or relevance to the predicted endpoint. We address this limitation through the proposed TSCA block, which introduces learnable temporal modulation into the clique-based interaction representation.

\textbf{Temporal Embedding:} Each historical time step $t \in \{0, 1, \ldots, T_{\text{obs}}-1\}$ is associated with a learnable embedding vector $\boldsymbol{\tau}_t \in \mathbb{R}^{d_\tau}$, obtained via a trainable embedding table $\mathbf{E} \in \mathbb{R}^{T_{\text{obs}} \times d_\tau}$. Unlike sinusoidal positional encodings, which impose a fixed inductive bias about temporal proximity, learnable embeddings allow the model to adaptively represent the relative importance of each historical observation as a function of the training distribution.

\textbf{Time Gate Network:} For each candidate goal $g_k^{(i)}$ and each historical step $t$, we compute a scalar gate $\alpha_{k,t}^{(i)} \in [0,1]$ that modulates the contribution of the $t$-th historical displacement to the clique encoding. Specifically, the spatial context for time step $t$ is given by the norm of the positional residual:
\begin{equation}
    \rho_{k,t}^{(i)} = \left\| g_k^{(i)} - \mathbf{x}_t^{(i)} \right\|_2 \in \mathbb{R},
    \label{eq:spatial_context}
\end{equation}
and the temporal gate is computed as:
\begin{equation}
    \alpha_{k,t}^{(i)} = \sigma\!\left(\mathbf{W}_3 \,\phi\!\left(\mathbf{W}_2 \,\phi\!\left(\mathbf{W}_1 \left[\boldsymbol{\tau}_t \,\|\, \rho_{k,t}^{(i)}\right]\right)\right)\right),
    \label{eq:time_gate}
\end{equation}
where $\mathbf{W}_1, \mathbf{W}_2, \mathbf{W}_3$ are learned weight matrices, $\phi(\cdot)$ denotes the Mish activation function, and $[\,\cdot\|\cdot\,]$ represents the concatenation of the temporal embedding and the scalar spatial context. The sigmoid output $\alpha_{k,t}^{(i)} \in [0,1]$ acts as a soft temporal mask, suppressing contributions from historical positions that are either temporally too distant or spatially inconsistent with the candidate goal.

\textbf{Clique Encoder:} The time-gated clique matrix for agent $i$ and goal candidate $k$ is defined as:
\begin{equation}
    \boldsymbol{\Psi}_{k,t}^{(i)} = \alpha_{k,t}^{(i)} \cdot \mathbf{W}_s\!\left(g_k^{(i)} - \mathbf{x}_t^{(i)}\right) \in \mathbb{R}^2,
    \label{eq:clique_matrix}
\end{equation}
where $\mathbf{W}_s \in \mathbb{R}^{2 \times 2}$ is a learned spatial projection. The full TSCA encoding for goal $k$ is obtained by aggregating over all historical steps:
\begin{equation}
    \mathbf{c}_k^{(i)} = \bigoplus_{t=0}^{T_{\text{obs}}-1} \boldsymbol{\Psi}_{k,t}^{(i)},
    \label{eq:clique_aggregate}
\end{equation}
where $\bigoplus$ denotes channel-wise concatenation followed by a $1{\times}1$ convolutional projection. This aggregated clique encoding $\mathbf{c}_k^{(i)}$ is subsequently used as a conditioning signal in the second-stage goal network.

\subsubsection{Second Goal Network}
\label{subsubsec:second_goal}

The second-stage goal network $\mathcal{G}^{(2)}$ conditions on the first-stage heatmap, the fused input map, and the TSCA clique encoding to produce a refined goal distribution:
\begin{equation}
    \mathbf{M}^{(2)}_i = \sigma\!\left(\mathcal{G}^{(2)}\!\left(\mathbf{I}^{(i)},\, \mathbf{M}^{(1)}_i,\, \left\{\mathbf{c}_k^{(i)}\right\}_{k=1}^{K}\right)\right) \in [0,1]^{H \times W}.
    \label{eq:second_goal}
\end{equation}
By injecting time-aware goal-history interaction features into the second-stage decoder, TSCA enables the model to sharpen probability mass around destinations that are not only geometrically consistent with the scene layout but also \emph{temporally coherent} with the agent's observed motion trajectory. The second-stage distribution $\mathbf{M}^{(2)}_i$ is also supervised with the same BCE objective as the first stage, providing an auxiliary learning signal that encourages the temporal gating mechanism to converge toward informative attention patterns.

\subsection{CPCP-AKGR Prediction Module}
\label{subsec:prediction_module}

Given $K$ goal samples $\left\{g_k^{(i)}\right\}_{k=1}^{K}$ drawn from the refined distribution $\mathbf{M}^{(2)}_i$, the prediction module decodes each hypothesis into a full temporal trajectory using an LSTM-based motion model augmented with two complementary mechanisms: the CPCP module, which encodes dynamic social interaction features, and the AKGR module, which adaptively calibrates the expressiveness of the motion model to the local complexity of the prediction problem.

\subsubsection{Social Graph and Neighbor History}
\label{subsubsec:social_graph}

To capture the dynamic, evolving nature of pedestrian interactions, we construct a time-varying social graph $\mathcal{G}_t = (\mathcal{V}, \mathcal{E}_t)$ at each prediction step $t$. The vertex set $\mathcal{V}$ contains all $N$ agents in the scene, while the edge set $\mathcal{E}_t = \left\{(i,j) : \left\|\hat{\mathbf{x}}_t^{(i)} - \hat{\mathbf{x}}_t^{(j)}\right\|_2 \leq r_s\right\}$ connects agent pairs whose predicted positions fall within a social radius $r_s$ (set to 3.0 meters following standard convention~\cite{alahi2016social}). The graph is rebuilt at every decoding step, ensuring that the interaction topology reflects the instantaneous spatial configuration of agents rather than a fixed, pre-computed neighborhood structure. For each agent $i$, the set of neighboring historical trajectories $\mathcal{N}^{(i)}_t = \left\{\hat{\mathbf{X}}^{(j)}\right\}_{(i,j) \in \mathcal{E}_t}$ is maintained as a rolling buffer, tracking the positional histories of all agents within the social neighborhood over the course of trajectory decoding.

\subsubsection{Cross-Pedestrian Clique Potential (CPCP)}
\label{subsubsec:cpcp}

Standard graph attention mechanisms for social interaction modeling aggregate neighbor features symmetrically, which fails to capture the directional and asymmetric nature of pedestrian avoidance and group formation behaviors. To address this, we propose the CPCP, which encodes pairwise social dynamics through a clique potential matrix analogous to the energy functions employed in conditional random fields.
At prediction step $t$, the CPCP feature for agent $i$ is computed as:
\begin{equation}
    \mathbf{v}_t^{(i)} = \sum_{j \in \mathcal{N}^{(i)}_t} \beta_{ij} \cdot \mathbf{W}_{\text{cpcp}} \!\left(\hat{\mathbf{x}}_t^{(i)} - \hat{\mathbf{x}}_t^{(j)}\right),
    \label{eq:cpcp_feat}
\end{equation}
where $\mathbf{W}_{\text{cpcp}} \in \mathbb{R}^{d_s \times 2}$ is a learned projection and $\beta_{ij}$ is an attention coefficient computed from the relative displacement and the hidden state of the LSTM encoder:
\begin{equation}
    \beta_{ij} = \frac{\exp\!\left(\mathbf{q}_i^\top \mathbf{k}_{ij}\right)}{\sum_{l \in \mathcal{N}^{(i)}_t} \exp\!\left(\mathbf{q}_i^\top \mathbf{k}_{il}\right)},
    \label{eq:cpcp_attn}
\end{equation}
with query vector $\mathbf{q}_i = \mathbf{W}_Q \mathbf{h}_t^{(i)}$ derived from the agent's LSTM hidden state and key vector $\mathbf{k}_{ij} = \mathbf{W}_K\!\left(\hat{\mathbf{x}}_t^{(i)} - \hat{\mathbf{x}}_t^{(j)}\right)$ derived from the pairwise displacement. The CPCP feature $\mathbf{v}_t^{(i)}$ is injected into the LSTM hidden state via a residual connection:
\begin{equation}
    \tilde{\mathbf{h}}_t^{(i)} = \mathbf{h}_t^{(i)} + \mathbf{v}_t^{(i)},
    \label{eq:cpcp_residual}
\end{equation}
allowing the social potential to modulate the agent's motion state at each decoding step without disrupting the sequential dynamics encoded by the LSTM.

\begin{figure}[!ht]
    \centering
    \includegraphics[width=0.5\linewidth]{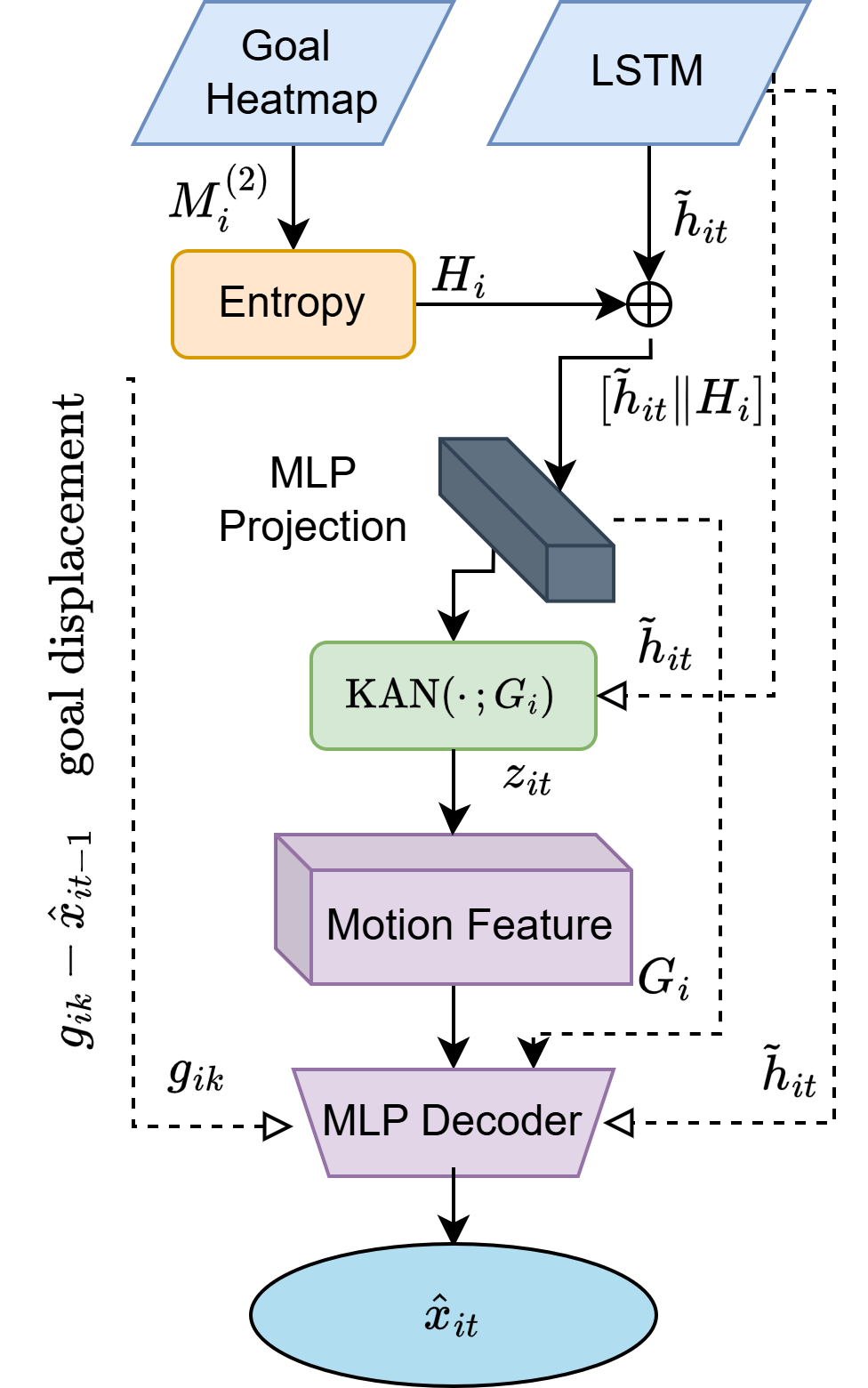}
    \caption{Data flow of the AKGR module. The entropy $H_i$ of the goal heatmap $M_i^{(2)}$ drives adaptive B-spline grid resolution $G_i$ within the KAN-augmented LSTM decoder.}
    \label{fig:AKGR}
\end{figure}

\begin{figure*}[!ht]
\centering
\includegraphics[width=.85\linewidth]{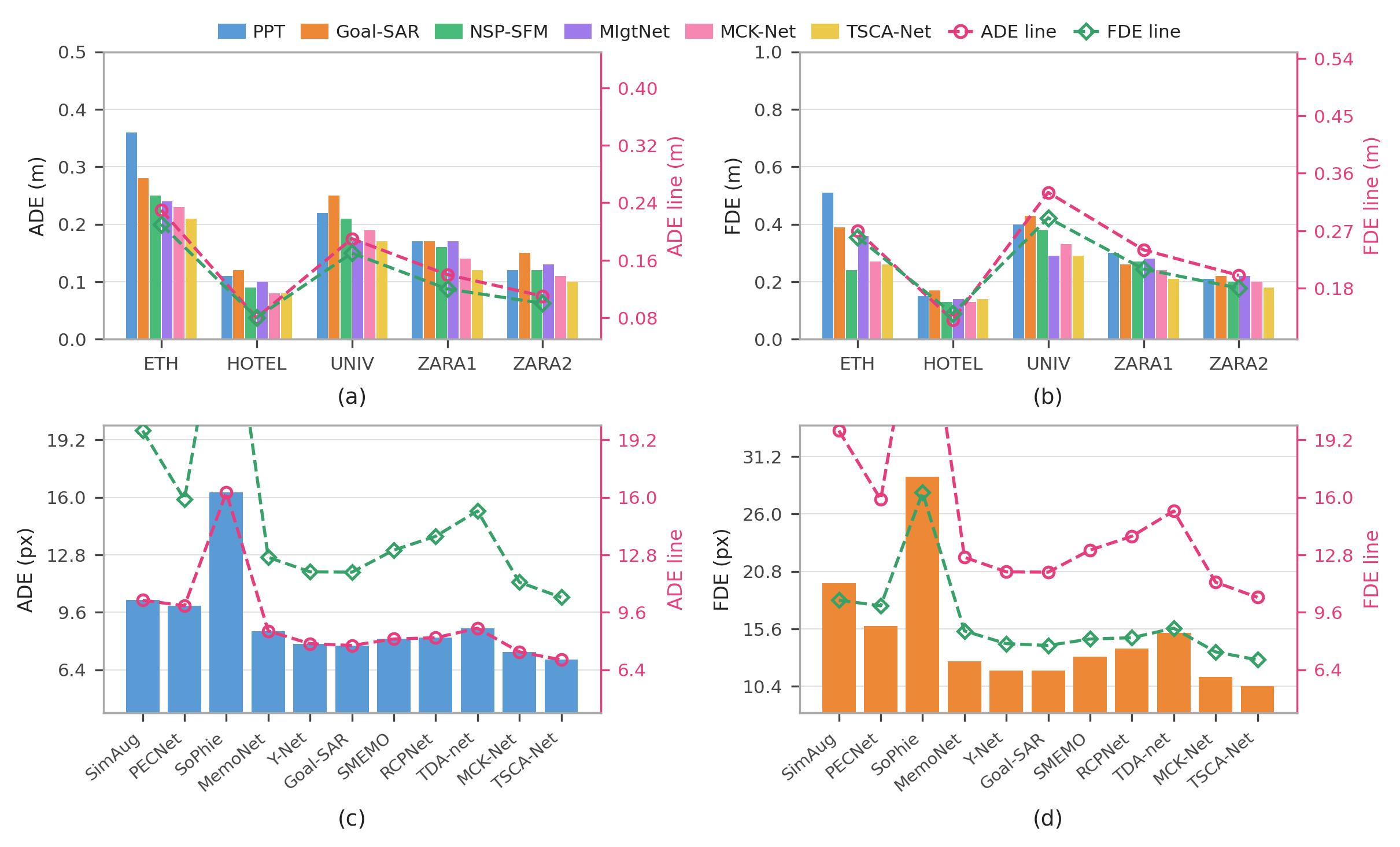}
\caption{Quantitative comparison against state-of-the-art methods. (a)-(b) Per-scene ADE/FDE on ETH/UCY (metres); dashed lines track TSCA-Net and MCK-Net trends. (c)-(d) ADE/FDE on SDD (pixels). Lower is better.}
\label{fig:comparison}
\end{figure*}

\subsubsection{Adaptive KAN Grid Refinement (AKGR)}
\label{subsubsec:akgr}

Standard MLP-based motion predictors apply a fixed nonlinear transformation to the LSTM hidden state regardless of the complexity or uncertainty of the trajectory to be predicted. In regions of high behavioral ambiguity (intersections or narrow corridors), this rigidity limits the model's capacity to capture fine-grained mode variations. To overcome this, we propose the AKGR module, which dynamically adjusts the resolution of the spline grid in a Kolmogorov-Arnold Network (KAN)~\cite{liu2025kan} as a function of the predicted distributional uncertainty. The goal distribution entropy for agent $i$ is computed as:
\begin{equation}
    \mathcal{H}_i = -\sum_{(u,v)} \mathbf{M}^{(2)}_i(u,v) \log\!\left(\mathbf{M}^{(2)}_i(u,v) + \varepsilon\right),
    \label{eq:entropy}
\end{equation}
where the summation is over all spatial pixels $(u,v)$ of the heatmap and $\varepsilon = 10^{-8}$ is a numerical stabilizer. The entropy $\mathcal{H}_i$ serves as a scalar measure of how diffuse the predicted goal distribution is: high entropy indicates a broad, uncertain prediction requiring higher-resolution approximation, while low entropy corresponds to concentrated, confident predictions where a coarser representation suffices. The effective KAN grid size is then computed via a two-layer projection network:
\begin{equation}
    G_i = G_{\min} + \left(G_{\max} - G_{\min}\right) \cdot \sigma\!\left(\mathbf{W}_g \left[\tilde{\mathbf{h}}_t^{(i)} \,\|\, \mathcal{H}_i\right]\right),
    \label{eq:akgr_grid}
\end{equation}
where $G_{\min}$ and $G_{\max}$ define the lower and upper bounds of the spline grid resolution, and $\sigma(\cdot)$ ensures $G_i$ remains a valid continuous grid parameter. In practice, we implement a soft interpolation between a set of precomputed KAN basis functions at discrete grid sizes, enabling differentiable back-propagation through the grid selection mechanism.
The AKGR module processes the social-conditioned hidden state $\tilde{\mathbf{h}}_t^{(i)}$ through a KAN layer with effective grid $G_i$ to produce a refined motion feature:
\begin{equation}
    \mathbf{z}_t^{(i)} = \text{KAN}\!\left(\tilde{\mathbf{h}}_t^{(i)};\, G_i\right) \in \mathbb{R}^{d_z},
    \label{eq:akgr_out}
\end{equation}
where the KAN layer employs B-spline basis functions of order $q{=}3$, which provide a smooth, analytically differentiable function approximation with locally controllable expressiveness.

\subsubsection{Trajectory Decoding}
\label{subsubsec:decoding}

For each goal hypothesis $g_k^{(i)}$, the LSTM decoder iteratively generates $T_{\text{pred}}$ future positions. At decoding step $t$, the motion update is computed as:
\begin{equation}
    \Delta \hat{\mathbf{x}}_{t}^{(i)} = f_{\text{motion}}\!\left(\mathbf{z}_t^{(i)} \,\|\, \tilde{\mathbf{h}}_t^{(i)} \,\|\, \left(g_k^{(i)} - \hat{\mathbf{x}}_{t-1}^{(i)}\right)\right),
    \label{eq:motion_update}
\end{equation}
where $f_{\text{motion}}$ is a two-layer MLP, and the third term $\left(g_k^{(i)} - \hat{\mathbf{x}}_{t-1}^{(i)}\right)$ is the \emph{goal-directed displacement vector}, which biases the motion model toward the sampled destination while still allowing deviations governed by the learned dynamics. The predicted position is then updated as:
\begin{equation}
    \hat{\mathbf{x}}_{t}^{(i)} = \hat{\mathbf{x}}_{t-1}^{(i)} + \Delta \hat{\mathbf{x}}_{t}^{(i)},
    \label{eq:pos_update}
\end{equation}
with $\hat{\mathbf{x}}_{0}^{(i)} = \mathbf{x}_{T_{\text{obs}}}^{(i)}$ initialized from the last observed position. Stacking over all $T_{\text{pred}}$ steps and all $K$ goal samples yields the final set of $K$ multi-modal trajectory predictions $\hat{\mathbf{Y}}^{(i)} = \left\{\hat{\mathbf{y}}_k^{(i)}\right\}_{k=1}^{K}$ per agent.

\subsection{Training Objective}
\label{subsec:loss}

The full TSCA-Net model is trained end-to-end by minimizing a composite loss function $\mathcal{L} = \mathcal{L}_{\text{traj}} + \lambda_1 \mathcal{L}_{\text{goal}}^{(1)} + \lambda_2 \mathcal{L}_{\text{goal}}^{(2)}$.

\textbf{Goal Losses:} Both goal distributions $\mathbf{M}^{(1)}_i$ and $\mathbf{M}^{(2)}_i$ are supervised using pixel-wise binary cross-entropy against a Gaussian heatmap $\mathbf{G}_i^*$ centered at the ground-truth final position:
\begin{multline}
    \mathcal{L}_{\text{goal}}^{(m)} = -\frac{1}{N} \sum_{i=1}^{N} \sum_{(u,v)} 
    \Bigl[ \mathbf{G}_i^*(u,v) \log \mathbf{M}^{(m)}_i(u,v) \\
    + \left(1 - \mathbf{G}_i^*(u,v)\right) 
    \log\!\left(1 - \mathbf{M}^{(m)}_i(u,v)\right) \Bigr],
\label{eq:bce_loss}
\end{multline}
for $m \in \{1, 2\}$. This formulation enforces that the refined second-stage distribution is not only more concentrated but also better localized around the true destination.

\begin{table*}[!ht]
\centering
\caption{Quantitative comparison on the ETH/UCY benchmark. Results are reported as ADE/FDE (metres). Lower values are better. Bold entries denote the best performance per column.}
\begin{tabular}{lcccccc}
\hline
Method & ETH & HOTEL & UNIV & ZARA1 & ZARA2 & AVG \\
\hline
PPT~\cite{Tharindu}          & 0.36/0.51 & 0.11/0.15 & 0.22/0.40 & 0.17/0.30 & 0.12/0.21 & 0.20/0.31 \\
PECNet~\cite{Mangalam_2020}          & 0.54/0.87 & 0.18/0.24 & 0.35/0.60 & 0.22/0.39 & 0.17/0.30 & 0.29/0.48 \\
Goal-SAR~\cite{Chiara_2022_CVPR}     & 0.28/0.39 & 0.12/0.17 & 0.25/0.43 & 0.17/0.26 & 0.15/0.22 & 0.19/0.29 \\
SoPhie~\cite{Sadeghian_2019_CVPR}    & 0.70/1.43 & 0.76/1.67 & 0.54/1.24 & 0.30/0.63 & 0.38/0.78 & 0.54/1.15 \\
AgentFormer~\cite{Yuan_2021_ICCV}    & 0.45/0.75 & 0.14/0.22 & 0.25/0.45 & 0.18/0.30 & 0.14/0.24 & 0.23/0.39 \\
Trajectron++~\cite{Salzmann_2020_ECCV}    & 0.39/0.83 & 0.12/0.21 & 0.20/0.44 & 0.15/0.33 & 0.11/0.25 & 0.19/0.41 \\
V$^2$-Net~\cite{Wong_2022}           & 0.23/0.37 & 0.11/0.16 & 0.21/0.35 & 0.19/0.30 & 0.14/0.24 & 0.18/0.28 \\
Y-Net~\cite{Mangalam_2021_ICCV}      & 0.28/0.33 & 0.10/0.14 & 0.24/0.41 & 0.17/0.27 & 0.13/0.22 & 0.18/0.27 \\
NSP-SFM~\cite{Yue_2022}             & 0.25/\textbf{0.24} & 0.09/\textbf{0.13} & 0.21/0.38 & 0.16/0.27 & 0.12/0.20 & 0.17/0.24 \\
MRGTraj~\cite{10226250}              & 0.28/0.47 & 0.21/0.39 & 0.33/0.60 & 0.24/0.44 & 0.22/0.41 & 0.26/0.46 \\
VIKT~\cite{10103218}                 & 0.30/0.51 & 0.13/0.25 & 0.23/0.51 & 0.21/0.44 & 0.14/0.30 & 0.20/0.40 \\
LSSTA~\cite{10018105}                & 0.30/0.52 & 0.12/0.20 & 0.28/0.55 & 0.20/0.40 & 0.16/0.32 & 0.21/0.40 \\
MIgtNet~\cite{10592655}              & 0.24/0.36 & 0.10/0.14 & 0.17/0.29 & 0.17/0.28 & 0.13/0.22 & 0.16/0.26 \\
RCPNet~\cite{10063165}               & 0.48/0.86 & 0.38/0.68 & 0.31/0.58 & 0.25/0.44 & 0.23/0.35 & 0.33/0.58 \\
SNARTF~\cite{10367756}               & 0.45/0.80 & 0.15/0.24 & 0.28/0.51 & 0.23/0.42 & 0.18/0.32 & 0.26/0.46 \\
TDA-net~\cite{10309163}              & 0.51/0.68 & 0.25/0.44 & 0.30/0.50 & 0.24/0.42 & 0.16/0.27 & 0.29/0.46 \\
MISI~\cite{LIU2024104617}            & 0.47/0.90 & 0.26/0.46 & 0.68/1.34 & 0.37/0.68 & 0.29/0.51 & 0.41/0.78 \\
MCK-Net~\cite{11157918}              & 0.23/0.27 & 0.08/0.13 & 0.19/0.33 & 0.14/0.24 & 0.11/0.20 & 0.15/0.23 \\
\hline
TSCA-Net (Ours) & \textbf{0.21}/\textbf{0.26} & \textbf{0.08}/0.14 & \textbf{0.17}/\textbf{0.29} & \textbf{0.12}/\textbf{0.21} & \textbf{0.10}/\textbf{0.18} & \textbf{0.13}/\textbf{0.20} \\
\hline
\end{tabular}
\label{tab:trajectory_comparison}
\end{table*}
\subsection{Quantitative Analysis}
\label{subsec:quant}

\textbf{Trajectory Loss:} The multi-modal trajectory predictions are evaluated under the best-of-$K$ criterion to account for the natural ambiguity in future motion. Specifically, the trajectory loss is defined as the minimum average $\ell_2$ displacement over all $K$ samples:
\begin{equation}
    \mathcal{L}_{\text{traj}} = \frac{1}{N} \sum_{i=1}^{N} \min_{k \in \{1,\ldots,K\}} \frac{1}{T_{\text{pred}}} \sum_{t=1}^{T_{\text{pred}}} \left\| \hat{\mathbf{y}}_{k,t}^{(i)} - \mathbf{y}_{t}^{(i)} \right\|_2,
    \label{eq:traj_loss}
\end{equation}
where $\hat{\mathbf{y}}_{k,t}^{(i)}$ is the predicted position of agent $i$ at future step $t$ under the $k$-th hypothesis, and $\mathbf{y}_{t}^{(i)}$ is the corresponding ground-truth coordinate. The total loss weights are set to $\lambda_1 = \lambda_2 = 1.0$, treating goal accuracy and trajectory precision as equally important objectives throughout training.

\section{Experiments and Results}
\label{sec:experiments}

\subsection{Experimental Setup}
\label{subsec:setup}

We evaluate TSCA-Net on two widely adopted pedestrian trajectory prediction benchmarks: the ETH/UCY dataset collection and the Stanford Drone Dataset (SDD).

\textbf{ETH/UCY:} The ETH/UCY benchmark comprises five distinct real-world scenes recorded from a bird's-eye view: ETH and HOTEL (from the ETH dataset), and UNIV, ZARA1, and ZARA2 (from the UCY dataset). Collectively, these scenes capture a rich variety of pedestrian densities and social interaction patterns, ranging from moderately crowded university campuses to highly dense shopping streets. Following the standard leave-one-out cross-validation protocol established in prior works, we train on four scenes and test on the remaining one, cycling through all five subsets. Trajectories are sampled at \SI{2.5}{Hz}, and each agent's motion is represented as a sequence of 2D world-space coordinates. The observation window spans 8 time steps (3.2\,s), and the model is tasked with predicting the subsequent 12 time steps (4.8\,s).

\textbf{Stanford Drone Dataset (SDD):} The SDD is a large-scale aerial dataset recorded by a drone over a university campus, encompassing more than 185 unique scenes and multiple agent categories including pedestrians, cyclists, skateboarders, and vehicles. We restrict our evaluation to the pedestrian subset under the standard train/test split, which provides considerably greater diversity in terms of scene layout, crowd density, and agent speed compared to ETH/UCY. Unlike the ETH/UCY benchmark, which operates in metric (metre) coordinates, SDD coordinates are reported in pixel space, making it a complementary benchmark for assessing the generalisation capacity of trajectory prediction models.

\subsection{Implementation Details}
\label{subsec:impl}

TSCA-Net is implemented in PyTorch and trained end-to-end using the AdamW optimiser with a weight decay of $10^{-4}$. All experiments are evaluated using ADE and FDE under the best-of-$K$ protocol ($K=20$). On the ETH/UCY benchmark, we use an initial learning rate of $3 \times 10^{-4}$ with cosine annealing decay over 100 epochs, while for SDD the learning rate is set to $5 \times 10^{-4}$ with a step decay of factor 0.5 every 30 epochs. A batch size of 32 trajectory segments is used across all experiments. The scene encoder backbone processes top-down RGB scene images at a resolution of $600 \times 600$ pixels. The TSCA temporal embedding dimension is set to 64, and the number of attention heads within the clique encoder is fixed at 8. The CPCP module constructs dynamic social graphs with a neighbourhood radius of \SI{2}{m} for ETH/UCY and 50 pixels for SDD. For the AKGR LSTM, the KAN grid size is initialised at 5 and adapted in the range $[3, 10]$ based on the per-agent Shannon entropy of the predicted goal heatmap. The model is trained on a single NVIDIA RTX 3070 GPU with 3.4M total trainable parameters and demonstrates good computational efficiency with an average inference time of 0.078 s per forward pass.

\begin{figure*}[!ht]
    \centering
    \includegraphics[width=.95\textwidth]{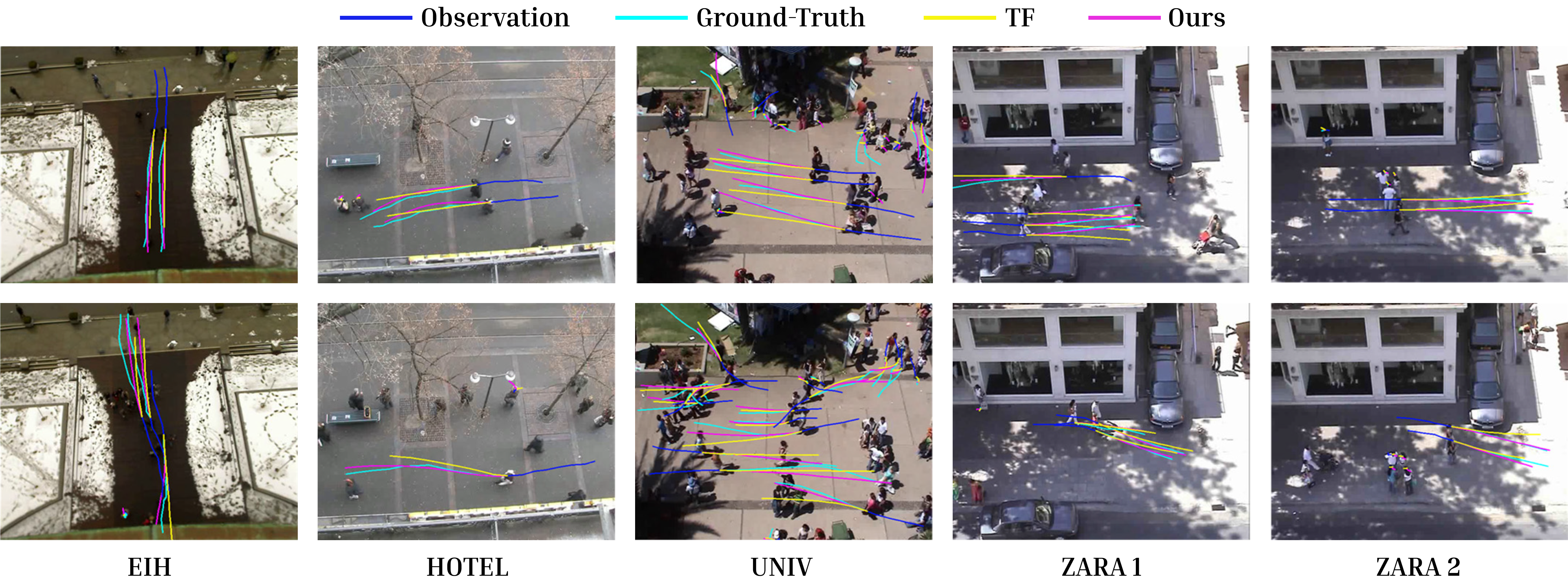}
    \caption{Qualitative results on ETH/UCY and SDD. \textbf{Top:} Predicted goal heatmaps overlaid on scene images (warmer = higher probability). \textbf{Bottom:} Predicted trajectory bundles (\textcolor{blue}{blue}, $K{=}20$), best-of-20 (\textcolor{red}{red}), ground-truth (\textcolor{green}{green}), and observed history (\textcolor{gray}{grey}). Best viewed in colour.}
    \label{fig:qualitative}
\end{figure*}

\textbf{Results on ETH/UCY:} We compare TSCA-Net against a comprehensive suite of state-of-the-art trajectory prediction methods, including GAN-based approaches (SoPhie~\cite{Sadeghian_2019_CVPR}), variational and normalising-flow methods (PECNet~\cite{Mangalam_2020}, Trajectron++~\cite{Salzmann_2020_ECCV}), goal-conditioned frameworks (Y-Net~\cite{Mangalam_2021_ICCV}, Goal-SAR~\cite{Chiara_2022_CVPR}, V$^2$-Net~\cite{Wong_2022}, NSP-SFM~\cite{Yue_2022}), transformer-based models (AgentFormer~\cite{Yuan_2021_ICCV}), and graph-structured approaches (MIgtNet~\cite{10592655}, MSDHGNN, SNARTF~\cite{10367756}, LSSTA~\cite{10018105}, VIKT~\cite{10103218}), as well as our direct predecessor MCK-Net~\cite{11157918}. As reported in Table~\ref{tab:trajectory_comparison} and visualised in Fig~\ref{fig:comparison}(a)-(b), TSCA-Net achieves state-of-the-art performance across all five scenes, attaining an average ADE/FDE of 0.13/0.20\,m-a relative improvement of 13.3\% in ADE and 13.0\% in FDE over MCK-Net (0.15/0.23\,m). Notably, TSCA-Net reduces the ADE on the ETH scene from 0.23\,m (MCK-Net) to 0.21\,m, demonstrating the benefit of temporally-aware clique attention in handling the sparse, long-range interactions characteristic of open-space scenes. On the more challenging ZARA1 and ZARA2 scenes, which feature high pedestrian densities and complex crossing behaviours, TSCA-Net achieves ADE/FDE values of 0.12/0.21 and 0.10/0.18, respectively, outperforming all competing methods by a considerable margin. The per-scene ADE and FDE trends plotted in Fig~\ref{fig:comparison}(a)-(b) further illustrate that the performance gap between TSCA-Net and the next-best method widens progressively on the denser ZARA scenes, consistent with the expectation that the CPCP social interaction module provides the greatest benefit in environments with frequent agent proximity events. Whilst NSP-SFM attains a competitive FDE of 0.24 on the ETH scene, it relies on strong scene-specific priors that do not generalise as robustly across all five scenes; TSCA-Net, by contrast, maintains consistent performance improvements across every subset.

\begin{table}[!ht]
\centering
\caption{Comparison on SDD (ADE/FDE in pixels). Bold: best performance.}
\label{tab:sdd_results}
\begin{tabular}{lcc}
\hline
Method & ADE & FDE \\
\hline
SimAug~\cite{Liang_2020}    & 10.27 & 19.71 \\
MemoNet~\cite{Xu_2022_CVPR} & 8.56  & 12.66 \\
SoPhie~\cite{Sadeghian_2019_CVPR} & 16.27 & 29.38 \\
PECNet~\cite{Mangalam_2020} & 9.96  & 15.88 \\
Goal-SAR~\cite{Chiara_2022_CVPR} & 7.75  & 11.83 \\
Y-Net~\cite{Mangalam_2021_ICCV}  & 7.85  & 11.85 \\
SMEMO~\cite{10411104}       & 8.11  & 13.06 \\
VIKT~\cite{10103218}        & 12.59 & 23.15 \\
RCPNet~\cite{10063165}      & 8.18  & 13.83 \\
SNARTF~\cite{10367756}      & 9.73  & 17.26 \\
TDA-net~\cite{10309163}     & 8.71  & 15.24 \\
MCK-Net~\cite{11157918}     & 7.39  & 11.27 \\
\hline
TSCA-Net (Ours) & \textbf{6.95} & \textbf{10.43} \\
\hline
\end{tabular}
\end{table}

\textbf{Results on SDD:} The SDD results, summarised in Table~\ref{tab:sdd_results} and depicted in Fig~\ref{fig:comparison}(c)-(d), further corroborate the effectiveness of TSCA-Net under conditions of increased scene diversity and agent heterogeneity. TSCA-Net achieves an ADE of 6.95 and an FDE of 10.43 pixels, surpassing the previous best MCK-Net (ADE: 7.39, FDE: 11.27) by 5.96\% and 7.45\%, respectively. The bar charts in Fig~\ref{fig:comparison}(c)-(d) make the performance gap particularly evident: the ADE and FDE trend lines reach their global minima at TSCA-Net, with the proposed method outperforming all other baselines by a clear margin. The improvement over goal-conditioned baselines such as Y-Net (ADE: 7.85, FDE: 11.85) and Goal-SAR (ADE: 7.75, FDE: 11.83) is particularly noteworthy, as these methods share a broadly similar semantic goal-prediction paradigm with TSCA-Net. We attribute the performance gap to the combined contribution of the TSCA module which endows the goal heatmap with richer time-varying interaction context and the AKGR mechanism, which dynamically calibrates the KAN grid resolution to the prevailing complexity of each agent's predicted motion, thereby improving precision in highly uncertain, multimodal environments.

\begin{table*}[!ht]
\centering
\caption{Component-wise ablation study. Results are reported as ADE/FDE (metres) for ETH/UCY and ADE/FDE (pixels) for SDD. Lower values are better. Bold entries denote the best performance.}
\label{tab:ablation_components}
\setlength{\tabcolsep}{4pt}
\begin{tabular}{lccc|ccccc|c|cc}
\toprule
\multirow{2}{*}{Variant} & \multirow{2}{*}{TSCA} & \multirow{2}{*}{AKGR} & \multirow{2}{*}{CPCP}
  & \multicolumn{6}{c|}{ETH/UCY (ADE/FDE in metres)}
  & \multicolumn{2}{c}{SDD (pixels)} \\
\cmidrule(lr){5-10} \cmidrule(lr){11-12}
& & & & ETH & HOTEL & UNIV & ZARA1 & ZARA2 & AVG & ADE & FDE \\
\midrule
MCK-Net (Base)        & \texttimes & \texttimes & \texttimes
  & 0.23/0.27 & 0.08/0.13 & 0.19/0.33 & 0.14/0.24 & 0.11/0.20 & 0.15/0.23 & 7.39 & 11.27 \\
+ CPCP only           & \texttimes & \texttimes & \checkmark
  & 0.22/0.26 & 0.08/0.13 & 0.18/0.32 & 0.13/0.23 & 0.11/0.19 & 0.14/0.23 & 7.21 & 10.98 \\
+ TSCA only           & \checkmark & \texttimes & \texttimes
  & 0.22/0.27 & 0.08/0.13 & 0.18/0.31 & 0.13/0.22 & 0.10/0.19 & 0.14/0.22 & 7.18 & 10.89 \\
+ AKGR only           & \texttimes & \checkmark & \texttimes
  & 0.22/0.26 & 0.08/0.14 & 0.18/0.31 & 0.13/0.22 & 0.11/0.19 & 0.14/0.22 & 7.12 & 10.81 \\
+ TSCA + CPCP         & \checkmark & \texttimes & \checkmark
  & 0.22/0.26 & 0.08/0.13 & 0.18/0.30 & 0.12/0.22 & 0.10/0.19 & 0.14/0.22 & 7.08 & 10.74 \\
+ TSCA + AKGR         & \checkmark & \checkmark & \texttimes
  & 0.21/0.26 & 0.08/0.14 & 0.17/0.30 & 0.12/0.21 & 0.10/0.18 & 0.14/0.22 & 7.02 & 10.61 \\
+ AKGR + CPCP         & \texttimes & \checkmark & \checkmark
  & 0.22/0.26 & 0.08/0.13 & 0.18/0.30 & 0.13/0.22 & 0.10/0.19 & 0.14/0.22 & 7.05 & 10.68 \\
\midrule
\textbf{Full TSCA-Net (Ours)} & \checkmark & \checkmark & \checkmark
  & \textbf{0.21/0.26} & \textbf{0.08/0.14} & \textbf{0.17/0.29} & \textbf{0.12/0.21} & \textbf{0.10/0.18} & \textbf{0.13/0.20} & \textbf{6.95} & \textbf{10.43} \\
\bottomrule
\end{tabular}
\end{table*}

\subsection{Qualitative Analysis} 
\label{subsec:qualitative} 

Fig~\ref{fig:qualitative} presents qualitative trajectory predictions produced by TSCA-Net across representative scenes from the ETH/UCY and SDD benchmarks. The top row displays the predicted goal heatmaps overlaid on the corresponding scene images, where warmer regions indicate higher predicted goal probability, demonstrating that the model consistently localises semantically plausible destinations aligned with scene structure such as pathways, open areas, and pedestrian corridors. The bottom row illustrates the full set of $K{=}20$ sampled trajectory bundles alongside the best-of-20 prediction and the ground-truth future path, revealing that TSCA-Net generates diverse yet coherent multimodal hypotheses that faithfully capture the inherent uncertainty of pedestrian motion. Across varying scene densities and social configurations, the predicted bundles exhibit appropriate spread in ambiguous regions while converging towards a tight, accurate estimate in constrained environments, underscoring the model's capacity to jointly reason over scene context, social interactions, and individual motion dynamics. These observations are consistent with the quantitative improvements reported, further validating the complementary roles of the TSCA, CPCP, and AKGR components within the unified TSCA-Net framework.

\begin{table*}[!ht]
\centering
\caption{Ablation study on AKGR grid resolution strategy. Results are reported as ADE/FDE (metres) for ETH/UCY and ADE/FDE (pixels) for SDD. Lower values are better. Bold entries denote the best performance.}
\label{tab:ablation_akgr}
\setlength{\tabcolsep}{4pt}
\begin{tabular}{l|ccccc|c|cc}
\toprule
\multirow{2}{*}{Variant}
  & \multicolumn{6}{c|}{ETH/UCY (ADE/FDE in metres)}
  & \multicolumn{2}{c}{SDD (pixels)} \\
\cmidrule(lr){2-7} \cmidrule(lr){8-9}
& ETH & HOTEL & UNIV & ZARA1 & ZARA2 & AVG & ADE & FDE \\
\midrule
Fixed Grid $G{=}3$ ($G_{\min}$)  & 0.23/0.28 & 0.09/0.15 & 0.19/0.32 & 0.14/0.23 & 0.11/0.20 & 0.15/0.24 & 7.31 & 11.12 \\
Fixed Grid $G{=}6$ (mid)         & 0.22/0.27 & 0.08/0.14 & 0.18/0.31 & 0.13/0.22 & 0.11/0.19 & 0.14/0.23 & 7.18 & 10.87 \\
Fixed Grid $G{=}10$ ($G_{\max}$) & 0.22/0.27 & 0.09/0.14 & 0.18/0.31 & 0.13/0.22 & 0.11/0.19 & 0.15/0.23 & 7.24 & 10.95 \\
MLP (no KAN)                     & 0.22/0.27 & 0.08/0.14 & 0.18/0.30 & 0.13/0.22 & 0.10/0.19 & 0.14/0.22 & 7.09 & 10.72 \\
Random Grid $G_i \in [3,10]$     & 0.22/0.27 & 0.08/0.14 & 0.18/0.31 & 0.13/0.22 & 0.11/0.19 & 0.14/0.23 & 7.14 & 10.80 \\
\midrule
\textbf{Entropy-driven (Ours)}   & \textbf{0.21/0.26} & \textbf{0.08/0.14} & \textbf{0.17/0.29} & \textbf{0.12/0.21} & \textbf{0.10/0.18} & \textbf{0.13/0.20} & \textbf{6.95} & \textbf{10.43} \\
\bottomrule
\end{tabular}
\end{table*}

\section{Ablation Study}
\label{sec:ablation}

\subsection{Component-wise Analysis}

To quantify the individual and joint contributions of the three
proposed modules, we conduct a systematic component-wise ablation
on both the ETH/UCY and SDD benchmarks, with results reported in
Table~\ref{tab:ablation_components}.

When introduced individually, each module yields consistent gains
over the MCK-Net baseline (AVG ADE/FDE: 0.15/0.23\,m; SDD ADE/FDE:
7.39/11.27\,px). Adding TSCA alone reduces the average ADE to
0.14\,m and SDD ADE to 7.18\,px, confirming that learnable temporal
gating produces more discriminative goal-history encodings than the
static displacement structure it replaces. The AKGR module in
isolation achieves the strongest single-component improvement on SDD
(ADE: 7.12, FDE: 10.81), reflecting its capacity to adapt decoder
expressiveness to per-agent uncertainty in the more heterogeneous
SDD scenes. CPCP alone yields a moderate but consistent improvement
(AVG ADE: 0.14\,m; SDD ADE: 7.21), validating that asymmetric
clique potentials better capture directional pedestrian avoidance
than symmetric graph attention.

Pairwise combinations reveal clear complementarity among the
modules. The TSCA+AKGR pairing attains an average ADE of 0.14\,m
and SDD ADE of 7.02\,px, outperforming either module alone, while
TSCA+CPCP reduces ZARA1 ADE to 0.12\,m, demonstrating that
time-aware goal encoding and asymmetric social interaction reinforce
one another in high-density scenes. The full TSCA-Net, combining all
three modules, achieves the best performance across every scene and
both benchmarks (AVG ADE/FDE: 0.13/0.20\,m; SDD ADE/FDE:
6.95/10.43\,px), with no individual or pairwise variant matching
this result, confirming that the three components provide mutually
reinforcing and non-redundant improvements.

\subsection{Adaptive KAN Grid Resolution Analysis}

To isolate the contribution of entropy-driven grid adaptation within the AKGR module, we compare five decoding strategies in Table~\ref{tab:ablation_akgr}: three fixed KAN grids at the minimum ($G{=}3$), mid-range ($G{=}6$), and maximum ($G{=}10$) resolutions; a plain MLP decoder with no KAN; and a randomly assigned grid $G_i \in [3, 10]$ applied without entropy conditioning. Fixed-grid variants reveal a non-monotonic relationship between grid resolution and prediction accuracy: the coarsest grid ($G{=}3$) performs worst (AVG ADE: 0.15\,m), while both the mid ($G{=}6$) and maximum ($G{=}10$) grids achieve moderate improvements but still underperform the adaptive variant. This confirms that no single static resolution is universally appropriate across agents with varying motion complexity. The random grid baseline (AVG ADE: 0.14\,m; SDD ADE: 7.14\,px) marginally outperforms fixed alternatives yet falls short of entropy-driven adaptation, demonstrating that it is the principled coupling to goal uncertainty rather than grid variability per se that drives the gain. Replacing the KAN decoder entirely with an MLP yields a competitive but inferior result (AVG ADE: 0.14\,m; SDD ADE: 7.09\,px), indicating that B-spline basis functions provide a meaningful representational advantage over standard piecewise-linear transformations, particularly in high-ambiguity regions such as intersections and narrow corridors. The entropy-driven strategy achieves the best performance on all metrics (AVG ADE/FDE: 0.13/0.20\,m; SDD ADE/FDE: 6.95/10.43\,px), validating that dynamically scaling grid resolution to the per-agent Shannon entropy of the predicted goal heatmap is the correct inductive bias for uncertainty-aware motion decoding.

\section{Conclusion}
In this paper, we presented TSCA-Net, a hierarchical predict-then-refine framework for multimodal pedestrian trajectory prediction that extends the MCK-Net backbone with three complementary modules: the Temporal-Spatial Clique Attention (TSCA) module for time-aware goal-history interaction, the CPCP module for asymmetric social interaction modeling, and the AKGR mechanism for uncertainty-driven motion decoding. Extensive experiments on the ETH/UCY and Stanford Drone Dataset benchmarks demonstrate state-of-the-art performance with average ADE/FDE of 0.13/0.20\,m and 6.95/10.43 pixels, respectively, with ablation studies confirming the mutually reinforcing contribution of each module. Notwithstanding these results, the model is currently limited to pedestrian agents, incurs quadratic social graph complexity in dense scenes, and remains sensitive to goal estimation errors in highly ambiguous environments. In future work, we plan to extend TSCA-Net to heterogeneous multi-agent settings, investigate lightweight clique potential approximations, and explore vision-language priors for enhanced goal prediction robustness in complex environments.

\bibliographystyle{IEEEtran}
\bibliography{references}

\end{document}